\documentclass{article}

\PassOptionsToPackage{numbers, compress}{natbib}
\usepackage[final]{neurips_2022}

\usepackage[utf8]{inputenc} % allow utf-8 input
\usepackage[T1]{fontenc}    % use 8-bit T1 fonts
\usepackage{hyperref}       % hyperlinks
\usepackage{url}            % simple URL typesetting
\usepackage{booktabs}       % professional-quality tables
\usepackage{amsfonts}       % blackboard math symbols
\usepackage{nicefrac}       % compact symbols for 1/2, etc.
\usepackage{microtype}      % microtypography
\usepackage{xcolor}         % colors

\usepackage{graphicx} % more modern
\usepackage{algorithm}
\usepackage{algorithmic}
\usepackage{caption}
\usepackage{mathtools}
\usepackage{stackengine} % eq align
\usepackage{bbm}
\usepackage{siunitx}
\usepackage{amsmath}
\usepackage{pifont}% 
\newcommand{\cmark}{\ding{51}}

\usepackage{mathrsfs} % \mathscr
\usepackage{makecell} % change line in table

\title{Semi-supervised Semantic Segmentation with Prototype-based Consistency Regularization}

\author{%
  Hai-Ming Xu\textsuperscript{1}~, Lingqiao Liu\textsuperscript{1}\thanks{Corresponding author}~, Qiuchen Bian\textsuperscript{2}~, Zhen Yang\textsuperscript{3} \\
  \textsuperscript{1}Australian Institute for Machine Learning, The University of Adelaide, \\ \textsuperscript{2}Northeastern University, \textsuperscript{3} Huawei Noah's Ark Lab\\
  \texttt{\{hai-ming.xu, lingqiao.liu\}@adelaide.edu.au}~,\\ \texttt{bian.qiu@northeastern.edu}~, \texttt{yang.zhen@hauwei.com}
}

\begin{document}

\maketitle

\begin{abstract}\label{sec:abstract}
Semi-supervised semantic segmentation requires the model to effectively propagate the label information from limited annotated images to unlabeled ones. A challenge for such a per-pixel prediction task is the large intra-class variation, i.e., regions belonging to the same class may exhibit a very different appearance even in the same picture. This diversity will make the label propagation hard from pixels to pixels. To address this problem, we propose a novel approach to regularize the distribution of within-class features to ease label propagation difficulty. Specifically, our approach encourages the consistency between the prediction from a linear predictor and the output from a prototype-based predictor, which implicitly encourages features from the same pseudo-class to be close to at least one within-class prototype while staying far from the other between-class prototypes. By further incorporating CutMix operations and a carefully-designed prototype maintenance strategy, we create a semi-supervised semantic segmentation algorithm that demonstrates superior performance over the state-of-the-art methods from extensive experimental evaluation on both Pascal VOC and Cityscapes benchmarks\footnote{Code is available at \url{https://github.com/HeimingX/semi_seg_proto}.}.
\end{abstract}

\section{Introduction}
Semantic segmentation is a fundamental task in computer vision and has been widely used in many vision applications~\cite{siam2018comparative,asgari2021deep,ouahabi2021deep}. Despite the advances, most existing successful semantic segmentation systems~\cite{long2015fully,chen2017deeplab,cheng2021per,zhou2022rethinking} are supervised, which require a large amount of annotated data, a time-consuming and costly process. Semi-supervised semantic segmentation~\cite{zou2020pseudoseg,zhang2020survey,ouali2020semi,ke2020guided,chen2021semi,zhong2021pixel,hu2021semi,wang2022semi} is a promising solution to this problem, which only requires a limited number of annotated images and aims to learn from both labeled and unlabeled data to improve the segmentation performance. Recent studies in semi-supervised learning approaches suggest that pseudo-labeling~\cite{Lee13Pseudolabel,arazo2020pseudo,zhang2021flexmatch} and consistency-based regularization~\cite{laine2016temporal,berthelot2019mixmatch,xie2020unsupervised} are two effective schemes to leverage the unlabeled data. Those two schemes are often integrated into a teacher-student learning paradigm: the teacher model generates pseudo labels to train a student model that takes a perturbed input~\cite{sohn2020fixmatch}. In such a scheme, and also for most pseudo-labeling-based approaches, the key to success is how to effectively propagate labels from the limited annotated images to the unlabeled ones. A challenge for
the semi-supervised semantic segmentation task
is the large intra-class variation, i.e., regions belonging to the same class may exhibit a very different appearance even in the same picture. This diversity will make the label propagation hard from pixels to pixels.

In this paper, we propose a novel approach to regularize the distribution of within-class features to ease label propagation difficulty. Our method adopts two segmentation heads (a.k.a, predictors): a standard linear predictor and a prototype-based predictor. The former has learnable parameters that could be updated through back-propagation, while the latter relies on a set of prototypes that are essentially local mean vectors and are calculated through running average. Our key idea is to encourage the consistency between the prediction from a linear predictor and the output from a prototype-based predictor. Such a scheme implicitly regularizes the feature representation: features from the same class must be close to at least one class prototype while staying far from the other class prototypes. We further incorporate CutMix operation~\cite{yun2019cutmix} to ensure such consistency is also preserved for perturbed (mixed) input images, which enhances the robustness of the feature representation. This gives rise to a new semi-supervised semantic segmentation algorithm that only involves one extra consistency loss to the state-of-the-art framework and can be readily plugged into other semi-supervised semantic segmentation methods. Despite its simplicity, it has demonstrated remarkable improvement over the baseline approach and competitive results compared to the state-of-the-art approaches, as discovered in our experimental study.

\section{Related Work}
\noindent \textbf{Semi-supervised Learning} has made great progress in recent years due to its economic learning philosophy~\cite{zhu2009introduction}. The success of most of the semi-supervised learning researches can attribute to the following two learning schemes: pseudo-labeling and consistency regularization. Pseudo-labeling based methods~\cite{Lee13Pseudolabel,cascante2020curriculum,arazo2020pseudo,zhang2021flexmatch} propose to train the model on unlabeled samples with pseudo labels generated from the up-to-date optimized model. While consistency regularization based methods~\cite{laine2016temporal,tarvainen2017mean,verma2019interpolation,berthelot2019mixmatch,xie2020unsupervised} build upon the \textit{smoothness assumption}~\cite{luo2018smooth} and encourage the model to perform consistent on the same example with different perturbations. The recently proposed semi-supervised method FixMatch~\cite{sohn2020fixmatch} successfully combine these two techniques together to produce the state-of-the-art classification performance. Our approach draws on the successful experience of general semi-supervised learning and applies it to the semi-supervised semantic segmentation task.

\noindent \textbf{Semi-supervised Semantic Segmentation} benefits from the development of general semi-supervised learning and various kinds of semi-supervised semantic segmentation algorithms have been proposed. For example, PseudoSeg method~\cite{zou2020pseudoseg} utilizes the Grad-CAM~\cite{selvaraju2017grad} trick to calibrate the generated pseudo-labels for semantic segmentation network training. While CPS~\cite{chen2021semi} builds two parallel networks to generate cross pseudo labels for each each.
CutMix-Seg method~\cite{french2019semi} introduces the CutMix augmentation into semantic segmentation to construct consistency constraints on unlabeled samples. Alternatively, CCT~\cite{ouali2020semi} chooses to insert perturbations into the manifold feature representation to enforce a consistent prediction. And U${}^2$PL~\cite{wang2022semi} proposes to make sufficient use of unreliable pseudo supervisions. Meanwhile, considering the class-imbalance problem of semi-supervised semantic segmentation, several researches~\cite{hu2021semi,he2021re,guan2022unbiased} have been published. Our approach is inspired by the observation that large intra-class variation hinders the label information propagation from pixels to pixels in semi-supervised semantic segmentation and we propose a prototype-based consistency regularization method to alleviate this problem which is novel for related literature.

\noindent \textbf{Prototype-based Learning} has been well studied in the machine learning area~\cite{hastie2009elements}. The nearest neighbors algorithm~\cite{cover1967nearest} is one of the earliest works to explore the use of prototypes. Recently, researchers have successfully used prototype-based learning to solve various problems, e.g., the prototypical networks~\cite{snell2017prototypical} for few-shot learning and prototype-based classifier for semantic segmentation~\cite{zhou2022rethinking}. Our work further introduces the prototype-based learning into the semi-supervised problem and proves its effectiveness.

\section{Our Approach}
In this section, we first give an overview of our approach and then introduce the core concept of prototype-based consistency regularization for semi-supervised semantic segmentation. Finally, we introduce how the prototype is constructed and maintained throughout the learning process. 

\begin{figure}[t]
\centering
\begin{center}
	\includegraphics[width=\linewidth]{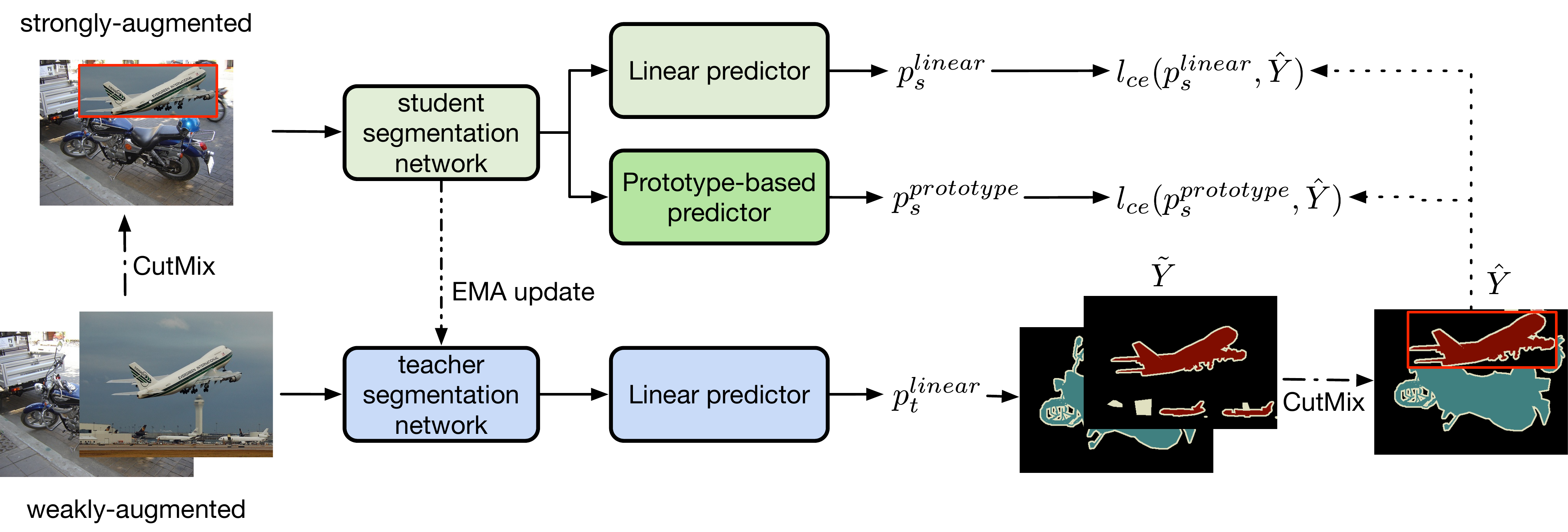}
\end{center}
\caption{Overview of our method. Our method is build upon the popular student-teacher frameworks with CutMix operations. In addition to the existing modules in such a framework, we further introduce a prototype-based predictor for the student model. The output $p_s^{prototype}$ of prototype-based predictor will be supervised with the pseudo-label generated from the linear predictor of teacher model. Such kind of consistency regularization will encourage the features from the same class to be closer than the features of other classes and ease the difficulty of propagating label information from pixels to pixels. This simple modification brings a significant improvement. 
}
\vspace{-0.25cm}
\label{fig:main_structure}
\end{figure}

\subsection{Preliminary}

\noindent\textbf{Problem setting:} Given a set of labeled training images $D^l = \{(I^l_i, Y^l_i)\}^{N_l}_{i=1}$ and a set of unlabeled images $D^u = \{I^u_i\}^{N_u}_{i=1}$, where $N_u \gg N_l$, semi-supervised semantic segmentation aims to learn a segmentation model from both the labeled and unlabeled images. We use $\tilde{Y}$ denote the segmentation output and $\tilde{Y}[a,b]$ indicates the output at the $(a,b)$ coordinate.

\noindent\textbf{Overview:} the overall structure of the proposed method is shown in Figure\ref{fig:main_structure},
our approach is built on top of the popular student-teacher framework for semi-supervised learning~\cite{tarvainen2017mean,sohn2020fixmatch,zhou2020time,nassar2021all,zhang2021flexmatch}. During the training procedure, the teacher model prediction will be selectively used as pseudo-labels for supervising the student model. In other words, the back-propagation is performed on the student model only. More specifically, the parameters of the teacher network are the exponential moving average of the student network parameters~\cite{tarvainen2017mean}. Following the common practice~\cite{sohn2020fixmatch}, we also adopt the weak-strong augmentation paradigm by feeding the teacher model weakly-augmented images and the student strongly-augmented images. In the context of image segmentation, we take the normal data augmentation (i.e., random crop and random horizontal flip of the input image) as the weak augmentation and CutMix~\cite{yun2019cutmix} as the strong data augmentation.

The key difference between our method and  existing methods~\cite{french2019semi,ouali2020semi,yang2021st++,chen2021semi,wang2022semi} is the use of both a linear predictor (in both teacher and student models) and a prototype-based predictor (in the student model only). As will be explained in the following section, the prediction from the teacher model's linear predictor will be used to create pseudo labels to supervise the training of the prototype-based predictor of student model. This process acts as a regularization that could benefit the label information propagation.

\subsection{Prototype-based Predictor for Semantic Segmentation}
Prototype-based classifier is a long-standing technique in machine learning~\cite{klawonn1999fuzzy,bobrowski1991c}. From its early form of the nearest neighbour classifier or the nearest mean classifier to prototypical networks in the few-shot learning literature~\cite{snell2017prototypical}, its idea of using prototypes instead of a parameterized classifier has been widely adopted in many fields. Very recently, prototype-based variety has been introduced into the semantic segmentation task~\cite{zhou2022rethinking} and has been proved to be effective under a fully-supervised setting. Formally, prototype-based classifier/predictors make the prediction by comparing test samples with a set of prototypes. The prototype can be a sample feature or the average of a set of sample features of the same class. Without loss of generality, we denote the prototype set as  $\mathcal{P}=\{(\mathsf{p}_i, y_i)\}$, with $\mathsf{p}_i$ indicate the prototype and $y_i$ is its associated class. Note that the number of prototypes could be larger than the number of classes. In other words, one class can have multiple prototypes for modelling its diversity. More formally, with the prototype set,
the classification decision can be made by using 
\begin{equation}
    \tilde{y} = y_k ~~s.t.~~ k = \mathop{\arg\max}\limits_i sim(x, \mathsf{p}_i), 
\end{equation}
where $sim(\cdot, \cdot)$ represents the similarity metric function, e.g., cosine distance. $\tilde{y}$ means the class assignment for the test data $x$. The posterior probability of assigning a sample to the $c$-th class can also be estimated in prototype-based classifier via:
\begin{align}~\label{eq:lab_proto_micro}
    p^{prototype}(y=c|x) = \frac{\exp  \biggl( \max_{i|y_i = c} sim (\mathsf{p}_i,x)/T\biggr)}{\sum_{t=1}^C \exp \biggl( \max_{j|y_j = t} sim(\mathsf{p}_j,x)/T\biggr)},
\end{align}where $T$ is the temperature parameter and can be empirically set. Note that Eq.~\ref{eq:lab_proto_micro} 
essentially uses the maximal similarity between a sample and prototypes of a class as the similarity between a sample and a class.

\subsection{Consistency Between Linear Predictor and Prototype-based Predictor}

Although both prototype-based classifiers and linear classifiers can be used for semantic segmentation~\cite{zhou2022rethinking}, they have quite different characteristics due to the nature of their decision-making process. Specifically, linear classifiers could allocate learnable parameters\footnote{Learnable parameters in the context means parameters that can be updated via back-propagation.} for each class, while prototype-based classifiers solely rely on a good feature representation such that samples from the same class will be close to at least one within-class prototypes while stay far from prototypes from other classes. Consequently, linear classifiers could leverage the learnable parameter to focus more on discriminative dimensions of a feature representation while suppressing irrelevant feature dimensions, i.e., by assigning a higher or lower weight to different dimensions. In contrast, prototype-based classifiers cannot leverage that and tend to require more discriminative feature representations.

The different characteristics of prototype-based and linear classifiers motivate us to design a loss to encourage the consistency of their predictions on unlabeled data to regularize the feature representation. Our key insight is that a good feature should support either type of classifier to make correct predictions. In addition to using two different types of classifiers, we also incorporate the CutMix~\cite{yun2019cutmix} strategy to enhance the above consistency regularization. CutMix augmentation is a popular ingredient in many state-of-the-art semi-supervised semantic segmentation methods~\cite{chen2021semi,liu2021perturbed,wang2022semi}. Specially, we first perform weak augmentation, e.g., random flip and crop operations, to the input images of the teacher model and obtain the pseudo-labels from the linear classifier. Next, we perform the CutMix operation by mixing two unlabeled images $mix(I_i, I_j)$ and their associated prediction $mix(\tilde{Y_i},\tilde{Y_j})$. The mixed image $mix(I_i, I_j)$ is fed to the student model and the output from the prototype-based classifier is then enforced to fit the pseudo-labels generated from $mix(\tilde{Y_i},\tilde{Y_j})$.  

\begin{algorithm}[t]
\caption{Global view of our approach}
\label{alg:global_view}
\begin{algorithmic}[1]
\REQUIRE ~~\\
 $D^l$: labeled set; \\
 $D^u$: unlabeled set; \\
 $T$: total number of epochs \\
\ENSURE ~~\\
 teacher semantic segmentation network with linear predictor only \\
 \hspace{-15pt}{\bf Process:}\\
 \STATE Prototype initialization, please refer to Algorithm~\ref{alg:proto_init} for details;\\
 \FOR{$\texttt{t} \gets [1 \to T]$}
    \STATE \textbf{Update student semantic segmentation network:} \\
    \STATE Sample $B$ examples from labeled set $D^l$ and unlabeled set $D^u$, respectively;\\
    \STATE \textit{For labeled data}, the student model is updated based on the given ground truth, please refer to Eq.(3)-(6) of main paper; \\
    \STATE \textit{For unlabeled data}, weakly augmented version is fed into the teacher model to generate pseudo-labels and the student model is updated with the strongly augmented unlabeled sample based on the pseudo-labels. Please refer to Eq. (8)-(10) of main paper; \\
    \STATE \textit{Update prototypes} based on the ground truth of labeled samples and the pseudo-labels of unlabeled samples, please refer to Eq. (11) of main paper; \\
    \STATE \textbf{Update teacher semantic segmentation network:} 
    \STATE exponential moving average (EMA) of the parameters of the student model.
\ENDFOR
\end{algorithmic}
\end{algorithm}

\noindent \textbf{Algorithm details:}
As a semi-supervised segmentation algorithm, we apply different loss functions for labeled images and unlabeled images.

For a batch of labeled images $\{(I^l_i, Y^l_i)\}_{i=1}^{\mathcal{B}^l}\in D^l$, we train both the linear predictor and the prototype-based predictor. The linear classifier $\{\mathbf{w}_i\}_{i=1}^{C}$ can produce a posterior probability estimation $p_s^{linear}(Y[a,b]=c|I^l_i)$
\begin{align}~\label{eq:lab_linear}
    p_s^{linear}(Y[a,b]=c|I^l_i) = \frac{\exp (\mathbf{w}_c^T \cdot \mathcal{F}_i^l[a,b] )}{\sum_{j=1}^{C} \exp (\mathbf{w}_j^T \cdot \mathcal{F}_i^l[a,b] )},
\end{align}where $\mathcal{F}_i^l[a,b] = f(A_0(I_i^l))$ means the feature extracted at location $(a,b)$ by first performing weak data augmentation $A_0$ to $I_i^l$ and then feed it to the feature extractor $f$. Meanwhile, the posterior probability of prototype-based predictor  $p_s^{prototype}(Y[a,b]=c|I^l_i)$ can also be estimated via Eq. \ref{eq:lab_proto_micro}.
We use cosine similarity for $sim(\cdot, \cdot)$ and empirically set the temperature hyperparameter $T$ to $0.1$. Based on the ground truth label $Y_i^l$, 
the student model will be optimized by the gradient back-propagated from the two predictors simultaneously
\begin{align}~\label{eq:lab_loss}
    & \mathcal{L}_l = \mathcal{L}_l^{linear} + \mathcal{L}_l^{prototype}, \ ~~~~~~\text{where} \\
    & \mathcal{L}_l^{linear} = \frac{1}{\mathcal{B}^l} \sum_{i}^{\mathcal{B}^l} l_{ce}\bigl(p_s^{linear}(Y|I_i^l), Y_i^l\bigr); \\
    & \mathcal{L}_l^{prototype} = \frac{1}{\mathcal{B}^l} \sum_{i}^{\mathcal{B}^l} l_{ce}\bigl(p_s^{prototype}(Y|I_i^l), Y_i^l\bigr).
\end{align}

For a batch of unlabeled images $\{I^u_i\}_{i=1}^{\mathcal{B}^u}\in D^u$, we first use the teacher model to estimate their posterior probability
\begin{align}~\label{eq:unlab_tea_linear}
    p_t^{linear}(Y[a,b]=c|I^u_i) = \frac{\exp (\mathbf{w}'_c{}^T \cdot \mathcal{F}_i^u[a,b] )}{\sum_{j=1}^{C} \exp (\mathbf{w}'_j{}^T \cdot \mathcal{F}_i^u[a,b])}
\end{align}where ${\{\mathbf{w}'_i\}}_{i=1}^C$ means the linear classifier weights of the teacher model and $\mathcal{F}_i^u[a,b] = f\bigl(A_0(I^u_i)\bigr)[a,b]$ denotes the extracted feature representation of pixel $(a,b)$ from a weakly augmented unlabeled images. Then, the class corresponding to the maximal posterior probability is the predicted class of a pixel in the given unlabeled sample, that is, $\tilde{Y}^u_i[a,b] = \arg \max_{c} p_t^{linear}(Y[a,b]=c|I^u_i)$. If $p_t^{linear}(\tilde{Y}[a,b]|I_i^u) \ge \tau$, where $\tau$ is a confidence threshold which is empirically set to $0.8$ in our study, $\tilde{Y}[a,b]$ will be used as pseudo-labels to train the student model. 

Meanwhile, for the student model we perform CutMix~\cite{yun2019cutmix} operation among weakly augmented unlabeled samples in the same batch to create an new image (essentially, the created mix-image can be considered as a strongly-augmented image), i.e., $\hat{I}^u_{ij} = mix\bigl(A_0(I^u_i), A_0(I^u_j)\bigr) \ s.t., \{i, j\} \in \mathcal{B}^u$, and their corresponding mixed prediction $\hat{Y}_{ij}^u = mix(\tilde{Y}^u_i, \tilde{Y}^u_j)$. Therefore, the student model can learn from the unlabeled samples through the following training objectives
\begin{align}~\label{eq:unlab_loss}
    & \mathcal{L}_u = \mathcal{L}_u^{linear} + \mathcal{L}_u^{prototype}, \ ~~~~~~\text{where} \\
    & \mathcal{L}_u^{linear} = \frac{1}{\mathcal{B}^u} \sum_{i,j\in \mathcal{B}^u} \sum_{(a,b)} l_{ce}\Bigl(p_s^{linear}\bigl(Y[a,b]|\hat{I}_{ij}^u\bigr), \ \hat{Y}_{ij}^u[a,b]\Bigr) \cdot \mathbbm{1}\Bigl(p_t^{linear}(\hat{Y}_{ij}^u[a,b]|\hat{I}_{ij}^u)\ge \tau\Bigr) \\
    & \mathcal{L}_u^{prototype} = \frac{1}{\mathcal{B}^u} \sum_{i,j\in \mathcal{B}^u} \sum_{(a,b)} l_{ce}\Bigl(p_s^{prototype}\bigl(Y[a,b]|\hat{I}_{ij}^u\bigr), \ \hat{Y}_{ij}^u[a,b]\Bigr) \cdot \mathbbm{1}\Bigl(p_t^{linear}(\hat{Y}_{ij}^u[a,b]|\hat{I}_{ij}^u)\ge \tau\Bigr)~\label{subeq:l_u_proto}
\end{align}where $p_s^{linear}(Y[a,b]|\hat{I}^u_{ij})$ and $p_s^{prototype}(Y[a,b]|\hat{I}^u_{ij})$ are posterior probability predictions from linear classifier and prototype-based classifier of student model respectively. Note that we use the student-teacher training for both the linear predictor and the prototype predictor, as shown in $\mathcal{L}_u^{linear}$ and $\mathcal{L}_u^{prototype}$ respectively. A global view of our approach is presented in Algorithm~\ref{alg:global_view}.

\noindent \textbf{Understand $\mathcal{L}_u^{prototype}$ in Eq.~\ref{subeq:l_u_proto}:} In order to better understand the proposed regularization loss term $\mathcal{L}_u^{prototype}$
% in our approach
, we can consider the following significantly-simplified version of our method by omitting the CutMix operation: now let's imagine at a certain point of the training process, the learned feature representation can successfully support the linear classifier in making a correct prediction for some pixels. This means there are at least some discriminative feature dimensions that can distinguish classes. Without loss of generality, let's assume the feature vector for each pixel consists of two parts $\mathbf{x} = [\mathbf{x}_d, \mathbf{x}_c]$, where $\mathbf{x}_d$ is the discriminative part while $\mathbf{x}_c$ is a less discriminative part, e.g., features shared by many classes. Linear classifiers can assign lower weights to $\mathbf{x}_c$ to suppress its impact, however, the impact of $\mathbf{x}_c$ cannot be avoided by using prototype-based classifiers. Thus from the supervision of the linear classifier, the training objective of optimizing the prototype-based classifier could further suppress the generation of $\mathbf{x}_c$. Geometrically, this also encourages the features from the same class gather around a finite set of prototypes and being apart from prototypes of other classes. In this way, the (pseudo) class label can propagate more easily from pixel to pixel, which in turn benefits the learning of the linear classifier.

\begin{algorithm}[t]
\caption{Prototype initialization}
\label{alg:proto_init}
\begin{algorithmic}[1]
\REQUIRE ~~\\
 $D^l$: labeled set \\
 $K$: number of prototypes per class \\
\ENSURE ~~\\
 initial prototypes \\
 \hspace{-15pt}{\bf Process:}\\
 \STATE \textbf{supervised training:} Train the semantic segmentation network on the subset of fully-labeled samples (please refer to Section~\ref{subsec:exp_setup} for training details);\\
 \STATE \textbf{feature extraction:} Use the trained segmentation network to extract feature representations of labeled samples (i.e. the feature representation before feed into the classifier of DeepLabv3+ and perform interpolation on the feature representation to match the input image size). We then sample a certain amount of pixels with their representations for each category; \\
\STATE \textbf{feature clustering:} Perform K-Means clustering (other clustering algorithms are also possible) on sampled pixel representations from each category. This step creates $K$ sub-classes for each category. We use the feature average of samples in each subclass to obtain the initial prototypes of each category.\\
\end{algorithmic}
\end{algorithm}

\subsection{Prototype Initialization and Update}\label{sec:multi_proto}
\noindent\textbf{Prototype initialization:}
The prototype-based classifier does not have learnable classifier parameters but relies on a set of good prototypes. Thus it is vitally important to carefully devise strategies to initialize and maintain the pool of prototypes. 

To initialize the prototypes,  we first use the given labeled samples to train the semantic segmentation network (with a linear predictor) in a fully-supervised way for several epochs. Then we extract pixel-wise feature representation for each class with the trained segmentation network. With the in-class pixel-wise feature representations, we propose to perform clustering on them to find out internal sub-classes, and the initial micro-prototypes will be obtained by averaging the feature representations within the same subclass. Please find the Algorithm~\ref{alg:proto_init} for prototype initialization details.

\smallskip
\noindent \textbf{Prototype update:} In our approach, the prototypes are dynamically updated from the features extracted from the labeled images and those from unlabeled samples during the semi-supervised learning process. 

When a labeled image is sampled, we assign each pixel to a prototype based on two conditions: (1) the assigned prototype $\mathsf{p}_k$ should belong to the same class as the pixel. (2) $\mathsf{p}_k$ should be the most similar prototype among all other prototypes in the same class. Once the assignment is done, we update $\mathsf{p}_k$ via
\begin{align}
    \mathsf{p}_{k}^{new} = \alpha \cdot \mathsf{p}_{k}^{old} + (1-\alpha) \cdot \mathcal{F}[a, b],
\end{align}where $\mathcal{F}[a, b]$ is the feature representation for the pixel at $(a,b)$. $\alpha$ is a hyper-parameter controlling the prototype update speed. We set $\alpha = 0.99$ throughout our experiment.

For unlabeled images, the ground-truth class label for each pixel is unavailable, thus we use pseudo-label instead. Recall that the pseudo-label is generated when the prediction confidence is higher than a threshold. Thus, not every pixel will be used to update the prototype.

Also, since prototype-based classifier is only used for images after the CutMix~\cite{yun2019cutmix} operation. In our implementation, we use features extracted from the CutMix images to update the prototype rather than the original images. Empirically we find this could slightly improve the performance.

\section{Experiments}
\subsection{Experimental Setup}\label{subsec:exp_setup}
Our experiment setting follows the recently proposed state-of-the-art work U$^2$PL~\cite{wang2022semi} including the evaluation datasets, semantic segmentation networks and training schedules for a fair comparison~\footnote{\url{https://github.com/Haochen-Wang409/U2PL} (Apache 2.0 license)}.
Some experimental details are listed as follows

\smallskip
\noindent \textbf{Datasets:} PASCAL VOC 2012~\cite{everingham2010pascal} is designed for visual object class recognition. It contains twenty foreground object classes and one background class. The standard partition of the dataset
for training/validation/testing are 1,464/1,449/1,556 images, respectively.
In the semi-supervised semantic segmentation literature, some researches~\cite{chen2021semi,hu2021semi,yang2021st++,wang2022semi} also include the augmented set~\cite{hariharan2011semantic} for model training. This augmented set contains 9,118 images with coarse annotations. In the literature \cite{wang2022semi}, two ways of selecting the labeled data are considered: the \textit{classic} and the \textit{blender} setting. The former selects labeled data from the original 1,464 candidate labeled images while the latter selects among all the 10,582 images. We evaluate our method on both settings. 

Cityscapes~\cite{cordts2016cityscapes} is an urban scene understanding benchmark. The initial 30 semantic classes are re-mapped into 19 classes for the semantic segmentation task. The training, validation and testing set includes 2,975, 500 and 1,525 finely annotated images respectively. For both of these two datasets, four kinds of label partitions are considered: 1/16, 1/8, 1/4 and 1/2. In this paper, we compare all methods under the identical released label splits from U$^2$PL~\cite{wang2022semi} for a fair comparison.

\smallskip
\noindent \textbf{Evaluation:} We use single scale cropping for the evaluation of PASCAL VOC 2012 and slide window evaluation for Cityscapes for its high resolution. The mean of Intersection over Union (mIoU) is adopted as the evaluation metric. All numbers reported in this paper are measured on the validation set of these two datasets.

\smallskip
\noindent \textbf{Methods:} We compare our approach with several peer-reviewed semi-supervised segmentation algorithms: 
Mean Teacher (NeurIPS 2017)~\cite{tarvainen2017mean}, CutMix-Seg (BMVC 2020)~\cite{french2019semi}, PseudoSeg (ICLR 2020)~\cite{zou2020pseudoseg}, CCT (CVPR 2020)~\cite{ouali2020semi}, GCT (ECCV 2020)~\cite{ke2020guided}, CPS (CVPR 2021)~\cite{chen2021semi}, PC${}^2$ Seg(ICCV 2021)~\cite{zhong2021pixel}, AEL (NeurIPS 2021)~\cite{hu2021semi} and U${}^2$PL (CVPR 2022)~\cite{wang2022semi}. Meanwhile, performance of supervised only on labeled data is also reported for a reference baseline. To make a fair comparison, we conduct all experiments based on the same codebase released by the authors of U${}^2$PL~\cite{wang2022semi}.

\smallskip
\noindent \textbf{Implementation Details:} Following the common practice, we use ResNet-101~\cite{he2016deep} pre-trained on ImageNet~\cite{krizhevsky2012imagenet} as our backbone and DeepLabv3+~\cite{chen2018encoder} as the decoder. We take the default segmentation head as the pixel-level linear classifier. 
The feature representations for constructing the prototypes of our approach are extracted from the output of ASPP module~\cite{chen2017deeplab}. Our experiments were run on 8 * NVIDIA Tesla V100 GPUs (memory is 32G/GPU).

For both datasets, we adopt stochastic gradient descent (SGD) as the optimizer and set batch size to 16 for model optimization. While other training details are slightly different, e.g., PASCAL VOC 2012 is trained with initial learning rate \num{1.0e-03}, weight decay \num{1.0e-4} and 80 training epochs; while Cityscapes is trained with initial learning rate \num{1.0e-2}, weight decay \num{5.0e-4} and 200 training epochs. Meanwhile, we use the polynomial policy to dynamically decay the learning rate along the whole training: $lr=lr_{init} \cdot (1 - \frac{iter}{total iter})^{0.8}$. 

\begin{table}[t]%\small %\footnotesize
\centering
\caption{
Comparing results of state-of-the-art algorithms on \textbf{PASCAL VOC 2012} \texttt{val} set with mIoU (\%) $\uparrow$ metric. Methods are trained on the \textbf{\textit{classic}} setting, i.e., the labeled images are selected from the original VOC \texttt{train} set, which consists of $1,464$ samples in total.
}
\label{tab:voc_clean}
\begin{tabular}{l|ccccc}
\toprule
Method & 
1/16 (92) & 1/8 (183) & 1/4 (366) & 1/2 (732) & Full (1464) \\
\midrule
Supervised Only & 
45.77 & 54.92 & 65.88 & 71.69 &72.50 \\
\midrule
Mean Teacher~\cite{tarvainen2017mean} & 
51.72 & 58.93 & 63.86 & 69.51 & 70.96 \\
CutMix-Seg~\cite{french2019semi} & 
52.16 & 63.47 & 69.46 & 73.73 & 76.54 \\
PseudoSeg~\cite{zou2020pseudoseg} & 
57.60 & 65.50 & 69.14 & 72.41 & 73.23 \\
PC${}^2$Seg~\cite{zhong2021pixel} & 
57.00 & 66.28 & 69.78 & 73.05 & 74.15 \\

U$^2$PL~\cite{wang2022semi} & 
67.98 & 69.15 & 73.66 & 76.16 & 79.49 \\

\midrule
\textbf{Ours} & 
\textbf{70.06} & 
\textbf{74.71} & 
\textbf{77.16} & 
\textbf{78.49} & 
\textbf{80.65} \\
\bottomrule
\end{tabular}
\end{table}
\begin{table}[t]
\centering
\caption{
Comparing results of state-of-the-art algorithms on \textbf{PASCAL VOC 2012} \texttt{val} set with mIoU (\%) $\uparrow$ metric. Methods are trained on the \textbf{\textit{blender}} setting, i.e., the labeled images are selected from the augmented VOC \texttt{train} set, which consists of $10,582$ samples in total.
}
\label{tab:voc_noisy}
\begin{tabular}{l | c c c c}
\toprule
Method & 
1/16 (662) & 1/8 (1323) & 1/4 (2646) & 1/2 (5291) \\
\midrule
Supervised Only & 
67.87 & 71.55 & 75.80 & 77.13 \\
\midrule
Mean Teacher~\cite{tarvainen2017mean} & 
70.51 & 71.53 & 73.02 & 76.58  \\
CutMix-Seg~\cite{french2019semi} & 
71.66 & 75.51 & 77.33 & 78.21  \\
CCT~\cite{ouali2020semi} & 
71.86 & 73.68 & 76.51 & 77.40 \\
GCT~\cite{ke2020guided} &
70.90 & 73.29 & 76.66 & 77.98 \\
CPS~\cite{chen2021semi} & 
74.48 & 76.44 & 77.68 & 78.64\\
AEL~\cite{hu2021semi} & 
77.20 & 77.57 & 78.06 & 80.29\\

U$^2$PL~\cite{wang2022semi} &
77.21 & 79.01 & 79.30 & 80.50 \\

\midrule
\textbf{Ours} & \textbf{78.60} & \textbf{80.71} & \textbf{80.78} & \textbf{80.91} \\

\bottomrule
\end{tabular}
\end{table}

\subsection{Comparison with State-of-the-Arts}
\smallskip
\noindent \textbf{Results on PASCAL VOC 2012 Dataset~\cite{everingham2010pascal}:} 
Table~\ref{tab:voc_clean} and Table~\ref{tab:voc_noisy} report the comparison
results
on PASCAL VOC 2012 validation set under different label quality settings.
First, the results in Table~\ref{tab:voc_clean} are obtained 
under the \textit{classic} setting and our approach achieves consistent performance improvements over the compared methods. Specifically, our method outperforms the Supervised Only baseline by a large margin especially for the fewer data settings, e.g., \textbf{+24.29\%} for 1/16 and \textbf{+19.79\%} for 1/8 setting respectively. Meanwhile, our approach also successfully beats other semi-supervised methods. Taking the recently proposed state-of-the-art method U${}^2$PL~\cite{wang2022semi} as an example, the performance gain of our approach reaches to \textbf{+5.56\%} and \textbf{+3.50\%} mIoU improvements under 1/8 and 1/4 label partitions, respectively. 

Table~\ref{tab:voc_noisy} presents comparison results on the \textit{blender} setting. It is clear that our proposed method still achieves overall significant improvement over all other baselines. For example, our method excels to the Supervised Only baseline over \textbf{10\%} mIoU on the 1/16 split. Compared with previous well performed algorithms, e.g., AEL~\cite{hu2021semi} and U$^2$PL~\cite{wang2022semi}, our approach yields superior segmentation performance, e.g., \textbf{+1.39\%}, \textbf{+1.70\%} and \textbf{+1.48\%} on 1/16, 1/8 and 1/4 label partitions respectively. 

\smallskip
\noindent \textbf{Results on Cityscapes Dataset~\cite{cordts2016cityscapes}:} Table~\ref{tab:city} provides comparison results of our method against several existing algorithms on Cityscapes validation set. 
Compared to Supervised Only baseline, our method achieves a great performance improvement due to the make use of unlabeled data, e.g., under the 1/16 label partition, our approach surpasses Supervised Only baseline by \textbf{7.67\%}. 
Then, compared to the simple Mean Teacher~\cite{tarvainen2017mean} baseline, our approach also performs better in all cases.
Furthermore, our approach is superior than the state-of-the-art algorithm U${}^2$PL~\cite{wang2022semi}, e.g., \textbf{Ours} excels to U${}^2$PL by \textbf{3.11\%}, \textbf{1.94\%} and \textbf{1.93\%} under the 1/16, 1/8 and 1/4 label partition, respectively.

Note that our method performs slightly worse than AEL~\cite{hu2021semi} on the 1/16 label partition, it is because the class imbalance issue is more severe on this partition, and the AEL method, which is specially designed for handling the class imbalance problem, thus gains greater improvement. 
Since the purpose of this paper is to explore the new consistency loss to alleviate intra-class variation for the semi-supervised semantic segmentation task,
we do not explicitly consider measures to handle the label imbalance issue. Theoretically, the techniques for solving label imbalance issues can also be incorporated into our method for optimizing the overall performance.

\begin{table}[t]
\centering
\caption{
Comparing results of state-of-the-art algorithms on \textbf{Cityscapes} \texttt{val} set with mIoU (\%) $\uparrow$ metric. Methods are trained on identical label partitions and the labeled images are selected from the Cityscapes \texttt{train} set, which consists of $2,975$ samples in total.
}
\label{tab:city}
\begin{tabular}{l|c c c c}
\toprule
Method & 
1/16 (186) & 1/8 (372) & 1/4 (744) & 1/2 (1488) \\
\midrule
Supervised Only & 
65.74 & 72.53 & 74.43 & 77.83 \\
\midrule
Mean Teacher~\cite{tarvainen2017mean} & 
69.03 & 72.06 & 74.20 & 78.15 \\
CutMix-Seg~\cite{french2019semi} & 
67.06 & 71.83 & 76.36 & 78.25 \\
CCT~\cite{ouali2020semi} & 
69.32 & 74.12 & 75.99 & 78.10 \\
GCT~\cite{ke2020guided} &
66.75 & 72.66 & 76.11 & 78.34 \\
CPS~\cite{chen2021semi} &
69.78 & 74.31 & 74.58 & 76.81 \\
AEL~\cite{hu2021semi} &
\textbf{74.45} & 75.55 & 77.48 & 79.01 \\
U$^2$PL~\cite{wang2022semi} & 70.30 & 74.37 & 76.47 & 79.05 \\

\midrule
\textbf{Ours} & 73.41 & \textbf{76.31} & \textbf{78.40} & \textbf{79.11} \\
\bottomrule
\end{tabular}
\end{table}

\begin{figure}[t]
\centering
\begin{center}
	\includegraphics[width=0.95\linewidth]{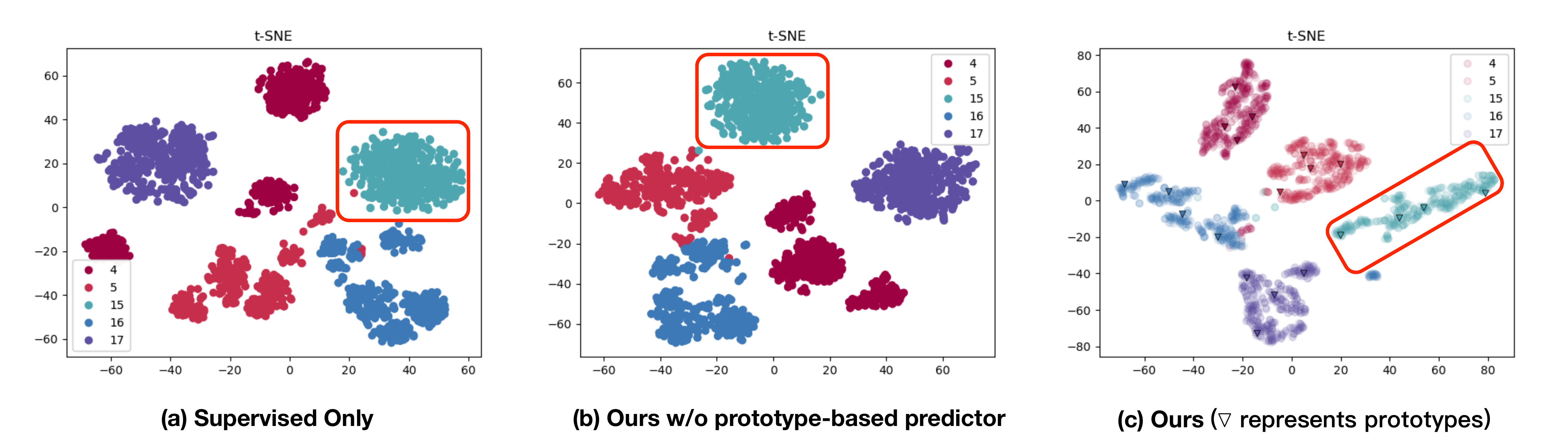}
\end{center}
\caption{Feature embedding visualizations of (a) Supervised Only, (b) ours without prototype-based predictor and (c) our method on the 1/16 partition of Pascal VOC 2012 using t-SNE~\cite{van2008visualizing}. As the data distribution shown in the red boxes, within-class feature representation of our method is more compact than the ones of the Supervised Only baseline and that of the variant without prototype-based predictor, which thus alleviates the large intra-class variation problem and eases the label information propagation from pixels to pixels. The corresponding relationship between the displayed category ID and semantic category is: \{4: ``boat'', 5: ``bottle'', 15: ``person'', 16: ``pottedplant'', 17: ``sheep''\}.
}
\label{fig:tsne}
\end{figure}

\subsection{Ablation Study}
\begin{figure}
    \begin{minipage}{.43\linewidth}
    \centering
    \includegraphics[width=0.8\linewidth]{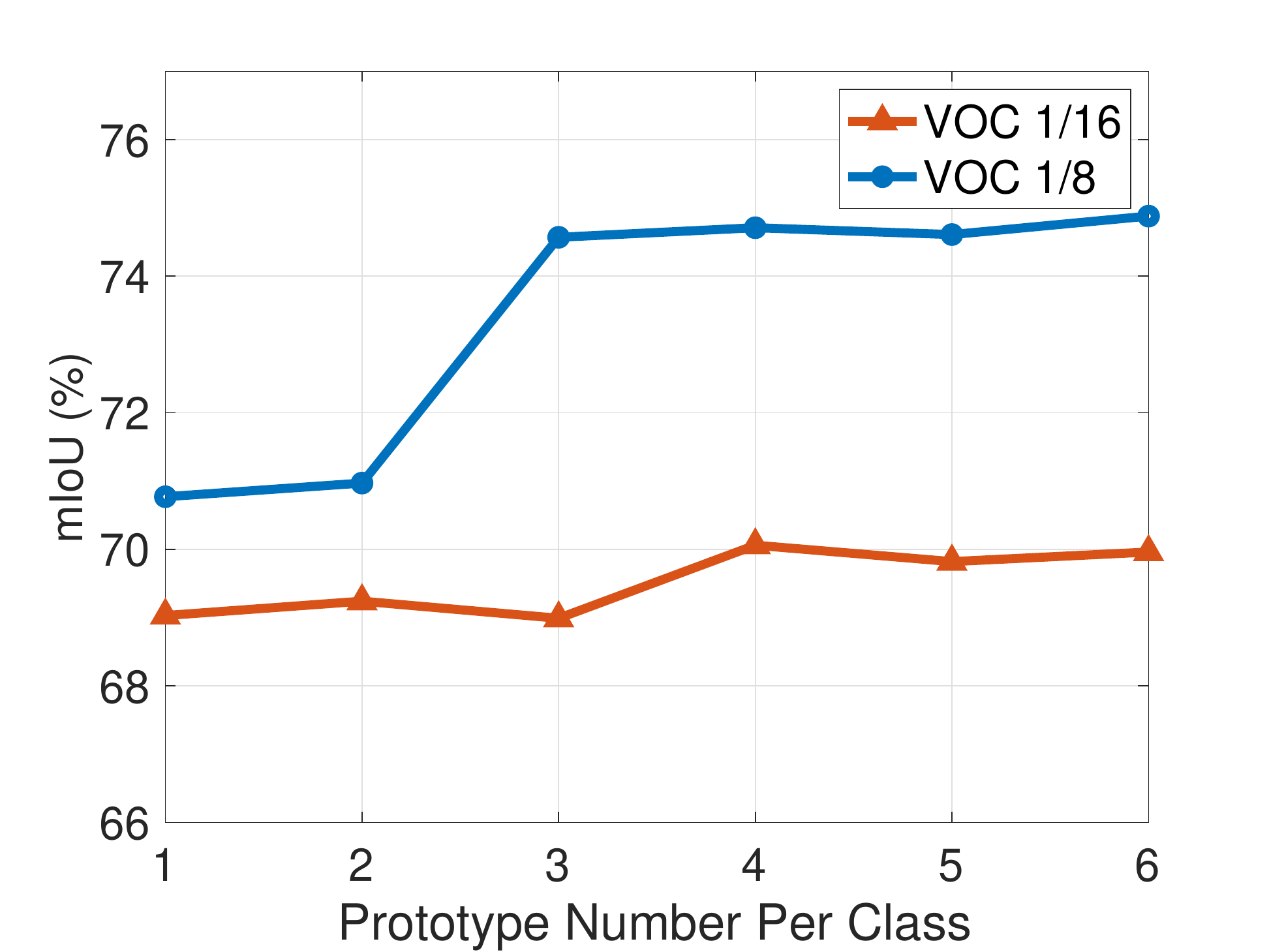}
    \caption{
    Abl. stu. number of prototype.
    }
    \label{fig:abl_proto_num}
    \end{minipage}%
    \begin{minipage}{.55\linewidth}
    \centering
    \captionof{table}{Ablation study on the effectiveness of different components of our approach.
    }
    \label{tab:abl_component}
    \begin{tabular}{l | c | c | c | c c }
         \toprule
         & \makecell[c]{linear \\ pred.} & \makecell[c]{proto. \\ pred.} & \makecell[c]{update \\ proto.} & 1/16 & 1/8 \\
         \midrule
         \ding{172} & \cmark & & & 67.95 & 70.99 \\
         \ding{173} & & \cmark & \cmark & 65.15 & 66.10 \\
         \ding{174} & \cmark & \cmark & & 67.53 & 71.89 \\
         \ding{175} & \cmark & \cmark & \cmark & \textbf{70.06} & \textbf{74.71} \\
         \bottomrule
    \end{tabular}
    \end{minipage}
\end{figure}

To investigate how our approach works on the semi-supervised semantic segmentation task, we conduct ablation studies on the \textit{classic} PASCAL VOC 2012 setting under 1/16 and 1/8 partition protocols from the following perspective views:

\smallskip
\noindent \textbf{Effectiveness of different components}: Table~\ref{tab:abl_component} presents ablation studies of several variants of our approach based on the usage of different components. The variant \ding{175}, which uses of all components, is the default setting of our approach and is presented here for a reference. The variant \ding{172} only contains a linear predictor and the prototype-based predictor is omitted. It is clear that the performance of this variant drops a lot compared to our approach and this proves that the prototype-based predictor plays a key role in our approach. On the contrary, the variant \ding{173} only maintains a prototype-based predictor and dynamically updates the prototypes during the training. The corresponding results are shown to be the worst among all the compared variants in Table~\ref{tab:abl_component}. We postulate the potential reason is that the prototype-based predictor itself is not good enough to generate high quality pseudo-labels without the help of the linear classifier under the limited labeled data setting and thus cannot fully leverage the large amount of unlabeled samples. The variant \ding{174} ablates the necessity of prototype update in our approach and the performance gap between this variant and variant \ding{175} shows that our approach will benefit from the prototype update procedure and produce overall best performance.

\smallskip
\noindent \textbf{Distribution of feature representation}:
The core idea of introducing prototype-based predictor in our approach is to utilize the prototype-based consistency regularization for alleviating the strong intra-class variation problem in semi-supervised semantic segmentation. Therefore, we are interested in the influence of our method on feature distribution. Figure~\ref{fig:tsne} presents the feature distribution of various methods for some classes of Pascal VOC 2012. We can find that our method tends to produce more compact feature distributions than other compared methods for every semantic class and such compact feature will ease the label information propagation from pixels to pixels and thus weaken the influence of intra-class variation. 

\smallskip
\noindent \textbf{Number of prototype}\label{subsec:proto_num}:
For the prototype-based classifier, the number of prototype is not restricted to be equal to the number of classes. In our approach, we construct multiple prototypes for each semantic class to handle the intra-class variation problem of semi-supervised semantic segmentation task. In order to explore the influence of the number of prototypes on our method, we conduct ablation studies on our approach with different number of prototypes. As the results shown in Figure~\ref{fig:abl_proto_num}, the performance is tend to be saturate when the prototype number reaches to 4 for each semantic class. Therefore, we empirically take this number as the default value of our approach.

\section{Limitations}\label{sec:limitation}
One underlying assumption about our approach is that we mainly consider convolutional based semantic segmentation networks.
Recently transformer-based algorithms~\cite{cheng2021per,xie2021segformer} are being investigated for semantic segmentation that are not explored in this paper and is left for future work.
One underlying assumption about our approach is that we mainly consider semantic segmentation networks of per-pixel prediction style.

\section{Conclusion}
Semi-supervised semantic segmentation aims to propagate label information from pixels to pixels effectively, but the large intra-class variation hinders the propagation ability. In this paper, we introduce a prototype-based predictor into our semi-supervised semantic segmentation network and propose a novel prototype-based consistency loss to regularize the intra-class feature representation to be more compact. Experimental results show that our method successfully achieves superior performance than other approaches.

\section{Impacts and Ethics}\label{sec:ethics}
This paper proposes a method for semi-supervised semantic segmentation which is a fundamental research topic in computer vision area and no potential negative societal impacts are known up to now.
In terms of ethics, we do not see immediate concerns for the models we introduce and to the best of our knowledge no datasets were used that have known ethical issues.

\smallskip
\textbf{Acknowledgements} 
This work is partially supported by Centre of Augmented Reasoning of the University of Adelaide. Meanwhile, we would like to thank the anonymous reviewers for their insightful comments. We also gratefully acknowledge the support of MindSpore\footnote{\url{https://www.mindspore.cn/}}, CANN (Computer Architecture for Neural
Networks) and Ascend AI Processor used for this research.

% \section*{References}
\clearpage
\bibliographystyle{plainnat}
\bibliography{egbib}

\begin{thebibliography}{51}
\providecommand{\natexlab}[1]{#1}
\providecommand{\url}[1]{\texttt{#1}}
\expandafter\ifx\csname urlstyle\endcsname\relax
  \providecommand{\doi}[1]{doi: #1}\else
  \providecommand{\doi}{doi: \begingroup \urlstyle{rm}\Url}\fi

\bibitem[Arazo et~al.(2020)Arazo, Ortego, Albert, O’Connor, and
  McGuinness]{arazo2020pseudo}
Eric Arazo, Diego Ortego, Paul Albert, Noel~E O’Connor, and Kevin McGuinness.
\newblock Pseudo-labeling and confirmation bias in deep semi-supervised
  learning.
\newblock In \emph{2020 International Joint Conference on Neural Networks
  (IJCNN)}, pages 1--8. IEEE, 2020.

\bibitem[Asgari~Taghanaki et~al.(2021)Asgari~Taghanaki, Abhishek, Cohen,
  Cohen-Adad, and Hamarneh]{asgari2021deep}
Saeid Asgari~Taghanaki, Kumar Abhishek, Joseph~Paul Cohen, Julien Cohen-Adad,
  and Ghassan Hamarneh.
\newblock Deep semantic segmentation of natural and medical images: a review.
\newblock \emph{Artificial Intelligence Review}, 54\penalty0 (1):\penalty0
  137--178, 2021.

\bibitem[Berthelot et~al.(2019)Berthelot, Carlini, Goodfellow, Papernot,
  Oliver, and Raffel]{berthelot2019mixmatch}
David Berthelot, Nicholas Carlini, Ian~J. Goodfellow, Nicolas Papernot, Avital
  Oliver, and Colin Raffel.
\newblock Mixmatch: {A} holistic approach to semi-supervised learning.
\newblock In Hanna~M. Wallach, Hugo Larochelle, Alina Beygelzimer, Florence
  d'Alch{\'{e}}{-}Buc, Emily~B. Fox, and Roman Garnett, editors, \emph{Advances
  in Neural Information Processing Systems 32: Annual Conference on Neural
  Information Processing Systems 2019, NeurIPS 2019, December 8-14, 2019,
  Vancouver, BC, Canada}, pages 5050--5060, 2019.
\newblock URL
  \url{https://proceedings.neurips.cc/paper/2019/hash/1cd138d0499a68f4bb72bee04bbec2d7-Abstract.html}.

\bibitem[Bobrowski and Bezdek(1991)]{bobrowski1991c}
Leon Bobrowski and James~C Bezdek.
\newblock C-means clustering with the l/sub l/and l/sub infinity/norms.
\newblock \emph{IEEE Transactions on Systems, Man, and Cybernetics},
  21\penalty0 (3):\penalty0 545--554, 1991.

\bibitem[Cascante-Bonilla et~al.(2021)Cascante-Bonilla, Tan, Qi, and
  Ordonez]{cascante2020curriculum}
Paola Cascante-Bonilla, Fuwen Tan, Yanjun Qi, and Vicente Ordonez.
\newblock Curriculum labeling: Revisiting pseudo-labeling for semi-supervised
  learning.
\newblock In \emph{Proceedings of the AAAI Conference on Artificial
  Intelligence}, volume~35, pages 6912--6920, 2021.

\bibitem[Chen et~al.(2017)Chen, Papandreou, Kokkinos, Murphy, and
  Yuille]{chen2017deeplab}
Liang-Chieh Chen, George Papandreou, Iasonas Kokkinos, Kevin Murphy, and Alan~L
  Yuille.
\newblock Deeplab: Semantic image segmentation with deep convolutional nets,
  atrous convolution, and fully connected crfs.
\newblock \emph{IEEE transactions on pattern analysis and machine
  intelligence}, 40\penalty0 (4):\penalty0 834--848, 2017.

\bibitem[Chen et~al.(2018)Chen, Zhu, Papandreou, Schroff, and
  Adam]{chen2018encoder}
Liang-Chieh Chen, Yukun Zhu, George Papandreou, Florian Schroff, and Hartwig
  Adam.
\newblock Encoder-decoder with atrous separable convolution for semantic image
  segmentation.
\newblock In \emph{Proceedings of the European conference on computer vision
  (ECCV)}, pages 801--818, 2018.

\bibitem[Chen et~al.(2021)Chen, Yuan, Zeng, and Wang]{chen2021semi}
Xiaokang Chen, Yuhui Yuan, Gang Zeng, and Jingdong Wang.
\newblock Semi-supervised semantic segmentation with cross pseudo supervision.
\newblock In \emph{Proceedings of the IEEE/CVF Conference on Computer Vision
  and Pattern Recognition}, pages 2613--2622, 2021.

\bibitem[Cheng et~al.(2021)Cheng, Schwing, and Kirillov]{cheng2021per}
Bowen Cheng, Alex Schwing, and Alexander Kirillov.
\newblock Per-pixel classification is not all you need for semantic
  segmentation.
\newblock \emph{Advances in Neural Information Processing Systems}, 34, 2021.

\bibitem[Cordts et~al.(2016)Cordts, Omran, Ramos, Rehfeld, Enzweiler, Benenson,
  Franke, Roth, and Schiele]{cordts2016cityscapes}
Marius Cordts, Mohamed Omran, Sebastian Ramos, Timo Rehfeld, Markus Enzweiler,
  Rodrigo Benenson, Uwe Franke, Stefan Roth, and Bernt Schiele.
\newblock The cityscapes dataset for semantic urban scene understanding.
\newblock In \emph{Proceedings of the IEEE conference on computer vision and
  pattern recognition}, pages 3213--3223, 2016.

\bibitem[Cover and Hart(1967)]{cover1967nearest}
Thomas Cover and Peter Hart.
\newblock Nearest neighbor pattern classification.
\newblock \emph{IEEE transactions on information theory}, 13\penalty0
  (1):\penalty0 21--27, 1967.

\bibitem[DeVries and Taylor(2017)]{devries2017improved}
Terrance DeVries and Graham~W Taylor.
\newblock Improved regularization of convolutional neural networks with cutout.
\newblock \emph{arXiv preprint arXiv:1708.04552}, 2017.

\bibitem[Everingham et~al.(2010)Everingham, Van~Gool, Williams, Winn, and
  Zisserman]{everingham2010pascal}
Mark Everingham, Luc Van~Gool, Christopher~KI Williams, John Winn, and Andrew
  Zisserman.
\newblock The pascal visual object classes (voc) challenge.
\newblock \emph{International journal of computer vision}, 88\penalty0
  (2):\penalty0 303--338, 2010.

\bibitem[French et~al.(2019)French, Laine, Aila, Mackiewicz, and
  Finlayson]{french2019semi}
Geoff French, Samuli Laine, Timo Aila, Michal Mackiewicz, and Graham Finlayson.
\newblock Semi-supervised semantic segmentation needs strong, varied
  perturbations.
\newblock \emph{The 31st British Machine Vision Virtual Conference}, 2019.

\bibitem[Guan et~al.(2022)Guan, Huang, Xiao, and Lu]{guan2022unbiased}
Dayan Guan, Jiaxing Huang, Aoran Xiao, and Shijian Lu.
\newblock Unbiased subclass regularization for semi-supervised semantic
  segmentation.
\newblock In \emph{Proceedings of the IEEE/CVF Conference on Computer Vision
  and Pattern Recognition}, pages 9968--9978, 2022.

\bibitem[Hariharan et~al.(2011)Hariharan, Arbel{\'a}ez, Bourdev, Maji, and
  Malik]{hariharan2011semantic}
Bharath Hariharan, Pablo Arbel{\'a}ez, Lubomir Bourdev, Subhransu Maji, and
  Jitendra Malik.
\newblock Semantic contours from inverse detectors.
\newblock In \emph{2011 international conference on computer vision}, pages
  991--998. IEEE, 2011.

\bibitem[Hastie et~al.(2009)Hastie, Tibshirani, Friedman, and
  Friedman]{hastie2009elements}
Trevor Hastie, Robert Tibshirani, Jerome~H Friedman, and Jerome~H Friedman.
\newblock \emph{The elements of statistical learning: data mining, inference,
  and prediction}, volume~2.
\newblock Springer, 2009.

\bibitem[He et~al.(2016)He, Zhang, Ren, and Sun]{he2016deep}
Kaiming He, Xiangyu Zhang, Shaoqing Ren, and Jian Sun.
\newblock Deep residual learning for image recognition.
\newblock In \emph{Proceedings of the IEEE conference on computer vision and
  pattern recognition}, pages 770--778, 2016.

\bibitem[He et~al.(2021)He, Yang, and Qi]{he2021re}
Ruifei He, Jihan Yang, and Xiaojuan Qi.
\newblock Re-distributing biased pseudo labels for semi-supervised semantic
  segmentation: A baseline investigation.
\newblock In \emph{Proceedings of the IEEE/CVF International Conference on
  Computer Vision}, pages 6930--6940, 2021.

\bibitem[Hu et~al.(2021)Hu, Wei, Hu, Ye, Cui, and Wang]{hu2021semi}
Hanzhe Hu, Fangyun Wei, Han Hu, Qiwei Ye, Jinshi Cui, and Liwei Wang.
\newblock Semi-supervised semantic segmentation via adaptive equalization
  learning.
\newblock \emph{Advances in Neural Information Processing Systems}, 34, 2021.

\bibitem[Ke et~al.(2020)Ke, Qiu, Li, Yan, and Lau]{ke2020guided}
Zhanghan Ke, Di~Qiu, Kaican Li, Qiong Yan, and Rynson~WH Lau.
\newblock Guided collaborative training for pixel-wise semi-supervised
  learning.
\newblock In \emph{European conference on computer vision}, pages 429--445.
  Springer, 2020.

\bibitem[Klawonn and Keller(1999)]{klawonn1999fuzzy}
Frank Klawonn and Annette Keller.
\newblock Fuzzy clustering based on modified distance measures.
\newblock In \emph{International Symposium on Intelligent Data Analysis}, pages
  291--301. Springer, 1999.

\bibitem[Krizhevsky et~al.(2012)Krizhevsky, Sutskever, and
  Hinton]{krizhevsky2012imagenet}
Alex Krizhevsky, Ilya Sutskever, and Geoffrey~E Hinton.
\newblock Imagenet classification with deep convolutional neural networks.
\newblock \emph{Advances in neural information processing systems}, 25, 2012.

\bibitem[Laine and Aila(2017)]{laine2016temporal}
Samuli Laine and Timo Aila.
\newblock Temporal ensembling for semi-supervised learning.
\newblock In \emph{International Conference on Learning Representations}, 2017.
\newblock URL \url{https://openreview.net/pdf?id=BJ6oOfqge}.

\bibitem[Lee(2013)]{Lee13Pseudolabel}
Donghyun Lee.
\newblock Pseudo-label: The simple and efficient semi-supervised learning
  method for deep neural networks.
\newblock In \emph{ICML Workshop on Challenges in Representation Learning},
  2013.

\bibitem[Liu et~al.(2022)Liu, Tian, Chen, Liu, Belagiannis, and
  Carneiro]{liu2021perturbed}
Yuyuan Liu, Yu~Tian, Yuanhong Chen, Fengbei Liu, Vasileios Belagiannis, and
  Gustavo Carneiro.
\newblock Perturbed and strict mean teachers for semi-supervised semantic
  segmentation.
\newblock \emph{Proceedings of the IEEE/CVF Conference on Computer Vision and
  Pattern Recognition}, 2022.

\bibitem[Long et~al.(2015)Long, Shelhamer, and Darrell]{long2015fully}
Jonathan Long, Evan Shelhamer, and Trevor Darrell.
\newblock Fully convolutional networks for semantic segmentation.
\newblock In \emph{Proceedings of the IEEE conference on computer vision and
  pattern recognition}, pages 3431--3440, 2015.

\bibitem[Luo et~al.(2018)Luo, Zhu, Li, Ren, and Zhang]{luo2018smooth}
Yucen Luo, Jun Zhu, Mengxi Li, Yong Ren, and Bo~Zhang.
\newblock Smooth neighbors on teacher graphs for semi-supervised learning.
\newblock In \emph{Proceedings of the IEEE conference on computer vision and
  pattern recognition}, pages 8896--8905, 2018.

\bibitem[Nassar et~al.(2021)Nassar, Herath, Abbasnejad, Buntine, and
  Haffari]{nassar2021all}
Islam Nassar, Samitha Herath, Ehsan Abbasnejad, Wray Buntine, and Gholamreza
  Haffari.
\newblock All labels are not created equal: Enhancing semi-supervision via
  label grouping and co-training.
\newblock In \emph{Proceedings of the IEEE/CVF Conference on Computer Vision
  and Pattern Recognition}, pages 7241--7250, 2021.

\bibitem[Olsson et~al.(2021)Olsson, Tranheden, Pinto, and
  Svensson]{olsson2021classmix}
Viktor Olsson, Wilhelm Tranheden, Juliano Pinto, and Lennart Svensson.
\newblock Classmix: Segmentation-based data augmentation for semi-supervised
  learning.
\newblock In \emph{Proceedings of the IEEE/CVF Winter Conference on
  Applications of Computer Vision}, pages 1369--1378, 2021.

\bibitem[Ouahabi and Taleb-Ahmed(2021)]{ouahabi2021deep}
Abdeldjalil Ouahabi and Abdelmalik Taleb-Ahmed.
\newblock Deep learning for real-time semantic segmentation: Application in
  ultrasound imaging.
\newblock \emph{Pattern Recognition Letters}, 144:\penalty0 27--34, 2021.

\bibitem[Ouali et~al.(2020)Ouali, Hudelot, and Tami]{ouali2020semi}
Yassine Ouali, C{\'e}line Hudelot, and Myriam Tami.
\newblock Semi-supervised semantic segmentation with cross-consistency
  training.
\newblock In \emph{Proceedings of the IEEE/CVF Conference on Computer Vision
  and Pattern Recognition}, pages 12674--12684, 2020.

\bibitem[Selvaraju et~al.(2017)Selvaraju, Cogswell, Das, Vedantam, Parikh, and
  Batra]{selvaraju2017grad}
Ramprasaath~R Selvaraju, Michael Cogswell, Abhishek Das, Ramakrishna Vedantam,
  Devi Parikh, and Dhruv Batra.
\newblock Grad-cam: Visual explanations from deep networks via gradient-based
  localization.
\newblock In \emph{Proceedings of the IEEE international conference on computer
  vision}, pages 618--626, 2017.

\bibitem[Siam et~al.(2018)Siam, Gamal, Abdel-Razek, Yogamani, Jagersand, and
  Zhang]{siam2018comparative}
Mennatullah Siam, Mostafa Gamal, Moemen Abdel-Razek, Senthil Yogamani, Martin
  Jagersand, and Hong Zhang.
\newblock A comparative study of real-time semantic segmentation for autonomous
  driving.
\newblock In \emph{Proceedings of the IEEE conference on computer vision and
  pattern recognition workshops}, pages 587--597, 2018.

\bibitem[Snell et~al.(2017)Snell, Swersky, and Zemel]{snell2017prototypical}
Jake Snell, Kevin Swersky, and Richard Zemel.
\newblock Prototypical networks for few-shot learning.
\newblock \emph{Advances in neural information processing systems}, 30, 2017.

\bibitem[Sohn et~al.(2020)Sohn, Berthelot, Carlini, Zhang, Zhang, Raffel,
  Cubuk, Kurakin, and Li]{sohn2020fixmatch}
Kihyuk Sohn, David Berthelot, Nicholas Carlini, Zizhao Zhang, Han Zhang,
  Colin~A Raffel, Ekin~Dogus Cubuk, Alexey Kurakin, and Chun-Liang Li.
\newblock Fixmatch: Simplifying semi-supervised learning with consistency and
  confidence.
\newblock \emph{Advances in Neural Information Processing Systems},
  33:\penalty0 596--608, 2020.

\bibitem[Tarvainen and Valpola(2017)]{tarvainen2017mean}
Antti Tarvainen and Harri Valpola.
\newblock Mean teachers are better role models: Weight-averaged consistency
  targets improve semi-supervised deep learning results.
\newblock \emph{Advances in neural information processing systems}, 30, 2017.

\bibitem[Van~der Maaten and Hinton(2008)]{van2008visualizing}
Laurens Van~der Maaten and Geoffrey Hinton.
\newblock Visualizing data using t-sne.
\newblock \emph{Journal of machine learning research}, 9\penalty0 (11), 2008.

\bibitem[Verma et~al.(2019)Verma, Lamb, Kannala, Bengio, and
  Lopez-Paz]{verma2019interpolation}
Vikas Verma, Alex Lamb, Juho Kannala, Yoshua Bengio, and David Lopez-Paz.
\newblock Interpolation consistency training for semi-supervised learning.
\newblock In \emph{Proceedings of the Twenty-Eighth International Joint
  Conference on Artificial Intelligence, {IJCAI-19}}, pages 3635--3641.
  International Joint Conferences on Artificial Intelligence Organization, 7
  2019.
\newblock \doi{10.24963/ijcai.2019/504}.
\newblock URL \url{https://doi.org/10.24963/ijcai.2019/504}.

\bibitem[Wang et~al.(2022)Wang, Wang, Shen, Fei, Li, Jin, Wu, Zhao, and
  Le]{wang2022semi}
Yuchao Wang, Haochen Wang, Yujun Shen, Jingjing Fei, Wei Li, Guoqiang Jin,
  Liwei Wu, Rui Zhao, and Xinyi Le.
\newblock Semi-supervised semantic segmentation using unreliable pseudo-labels.
\newblock \emph{Proceedings of the IEEE/CVF Conference on Computer Vision and
  Pattern Recognition}, 2022.

\bibitem[Xie et~al.(2021)Xie, Wang, Yu, Anandkumar, Alvarez, and
  Luo]{xie2021segformer}
Enze Xie, Wenhai Wang, Zhiding Yu, Anima Anandkumar, Jose~M Alvarez, and Ping
  Luo.
\newblock Segformer: Simple and efficient design for semantic segmentation with
  transformers.
\newblock \emph{Advances in Neural Information Processing Systems}, 34, 2021.

\bibitem[Xie et~al.(2020)Xie, Dai, Hovy, Luong, and Le]{xie2020unsupervised}
Qizhe Xie, Zihang Dai, Eduard Hovy, Thang Luong, and Quoc Le.
\newblock Unsupervised data augmentation for consistency training.
\newblock \emph{Advances in Neural Information Processing Systems},
  33:\penalty0 6256--6268, 2020.

\bibitem[Yang et~al.(2022)Yang, Zhuo, Qi, Shi, and Gao]{yang2021st++}
Lihe Yang, Wei Zhuo, Lei Qi, Yinghuan Shi, and Yang Gao.
\newblock St++: Make self-training work better for semi-supervised semantic
  segmentation.
\newblock \emph{Proceedings of the IEEE/CVF Conference on Computer Vision and
  Pattern Recognition}, 2022.

\bibitem[Yun et~al.(2019)Yun, Han, Oh, Chun, Choe, and Yoo]{yun2019cutmix}
Sangdoo Yun, Dongyoon Han, Seong~Joon Oh, Sanghyuk Chun, Junsuk Choe, and
  Youngjoon Yoo.
\newblock Cutmix: Regularization strategy to train strong classifiers with
  localizable features.
\newblock In \emph{Proceedings of the IEEE/CVF international conference on
  computer vision}, pages 6023--6032, 2019.

\bibitem[Zhang et~al.(2021)Zhang, Wang, Hou, Wu, Wang, Okumura, and
  Shinozaki]{zhang2021flexmatch}
Bowen Zhang, Yidong Wang, Wenxin Hou, Hao Wu, Jindong Wang, Manabu Okumura, and
  Takahiro Shinozaki.
\newblock Flexmatch: Boosting semi-supervised learning with curriculum pseudo
  labeling.
\newblock \emph{Advances in Neural Information Processing Systems}, 34, 2021.

\bibitem[Zhang et~al.(2020)Zhang, Zhou, Zhao, Man, Liu, and
  Yao]{zhang2020survey}
Man Zhang, Yong Zhou, Jiaqi Zhao, Yiyun Man, Bing Liu, and Rui Yao.
\newblock A survey of semi-and weakly supervised semantic segmentation of
  images.
\newblock \emph{Artificial Intelligence Review}, 53\penalty0 (6):\penalty0
  4259--4288, 2020.

\bibitem[Zhong et~al.(2021)Zhong, Yuan, Wu, Yuan, Peng, and
  Wang]{zhong2021pixel}
Yuanyi Zhong, Bodi Yuan, Hong Wu, Zhiqiang Yuan, Jian Peng, and Yu-Xiong Wang.
\newblock Pixel contrastive-consistent semi-supervised semantic segmentation.
\newblock In \emph{Proceedings of the IEEE/CVF International Conference on
  Computer Vision}, pages 7273--7282, 2021.

\bibitem[Zhou et~al.(2022)Zhou, Wang, Konukoglu, and
  Van~Gool]{zhou2022rethinking}
Tianfei Zhou, Wenguan Wang, Ender Konukoglu, and Luc Van~Gool.
\newblock Rethinking semantic segmentation: A prototype view.
\newblock \emph{Proceedings of the IEEE/CVF Conference on Computer Vision and
  Pattern Recognition}, 2022.

\bibitem[Zhou et~al.(2020)Zhou, Wang, and Bilmes]{zhou2020time}
Tianyi Zhou, Shengjie Wang, and Jeff Bilmes.
\newblock Time-consistent self-supervision for semi-supervised learning.
\newblock In \emph{International Conference on Machine Learning}, pages
  11523--11533. PMLR, 2020.

\bibitem[Zhu and Goldberg(2009)]{zhu2009introduction}
Xiaojin Zhu and Andrew~B Goldberg.
\newblock Introduction to semi-supervised learning.
\newblock \emph{Synthesis lectures on artificial intelligence and machine
  learning}, 3\penalty0 (1):\penalty0 1--130, 2009.

\bibitem[Zou et~al.(2021)Zou, Zhang, Zhang, Li, Bian, Huang, and
  Pfister]{zou2020pseudoseg}
Yuliang Zou, Zizhao Zhang, Han Zhang, Chun-Liang Li, Xiao Bian, Jia-Bin Huang,
  and Tomas Pfister.
\newblock Pseudoseg: Designing pseudo labels for semantic segmentation.
\newblock In \emph{International Conference on Learning Representations}, 2021.
\newblock URL \url{https://openreview.net/forum?id=-TwO99rbVRu}.

\end{thebibliography}

%%%%%%%%%%%%%%%%%%%%%%%%%%%%%%%%%%%%%%%%%%%%%%%%%%%%%%%%%%%%
\section*{Checklist}

%%% BEGIN INSTRUCTIONS %%%
% The checklist follows the references.  Please
% read the checklist guidelines carefully for information on how to answer these
% questions.  For each question, change the default \answerTODO{} to \answerYes{},
% \answerNo{}, or \answerNA{}.  You are strongly encouraged to include a {\bf
% justification to your answer}, either by referencing the appropriate section of
% your paper or providing a brief inline description.  For example:
% \begin{itemize}
%   \item Did you include the license to the code and datasets? \answerYes{See Section~\ref{gen_inst}.}
%   \item Did you include the license to the code and datasets? \answerNo{The code and the data are proprietary.}
%   \item Did you include the license to the code and datasets? \answerNA{}
% \end{itemize}
% Please do not modify the questions and only use the provided macros for your
% answers.  Note that the Checklist section does not count towards the page
% limit.  In your paper, please delete this instructions block and only keep the
% Checklist section heading above along with the questions/answers below.
%%% END INSTRUCTIONS %%%

\begin{enumerate}

\item For all authors...
\begin{enumerate}
  \item Do the main claims made in the abstract and introduction accurately reflect the paper's contributions and scope?
    \answerYes{}
  \item Did you describe the limitations of your work?
    \answerYes{See Section~\ref{sec:limitation}}
  \item Did you discuss any potential negative societal impacts of your work?
    \answerYes{See Section~\ref{sec:ethics}}
  \item Have you read the ethics review guidelines and ensured that your paper conforms to them?
    \answerYes{See Section~\ref{sec:ethics}}
\end{enumerate}

\item If you are including theoretical results...
\begin{enumerate}
  \item Did you state the full set of assumptions of all theoretical results?
    % \answerTODO{}
    \answerNA{}
        \item Did you include complete proofs of all theoretical results?
    % \answerTODO{}
    \answerNA{}
\end{enumerate}

\item If you ran experiments...
\begin{enumerate}
  \item Did you include the code, data, and instructions needed to reproduce the main experimental results (either in the supplemental material or as a URL)?
    \answerYes{See Abstract section.}
  \item Did you specify all the training details (e.g., data splits, hyperparameters, how they were chosen)?
    % \answerTODO{}
    \answerYes{See Sections~\ref{subsec:exp_setup} and~\ref{subsec:proto_num}.}
        \item Did you report error bars (e.g., with respect to the random seed after running experiments multiple times)?
    % \answerTODO{}
    \answerNo{All compared algorithms are experimented on the same random seed for a fair comparison in our paper.}
        \item Did you include the total amount of compute and the type of resources used (e.g., type of GPUs, internal cluster, or cloud provider)?
    % \answerTODO{}
    \answerYes{See Sections~\ref{subsec:exp_setup}}
\end{enumerate}

\item If you are using existing assets (e.g., code, data, models) or curating/releasing new assets...
\begin{enumerate}
  \item If your work uses existing assets, did you cite the creators?
    % \answerTODO{}
    \answerYes{See Sections~\ref{subsec:exp_setup}}
  \item Did you mention the license of the assets?
    % \answerTODO{}
    \answerYes{See Sections~\ref{subsec:exp_setup}}
  \item Did you include any new assets either in the supplemental material or as a URL?
    % \answerTODO{}
    \answerNo{}
  \item Did you discuss whether and how consent was obtained from people whose data you're using/curating?
    % \answerTODO{}
    \answerNo{Data used in this paper has been open source for research purpose.}
  \item Did you discuss whether the data you are using/curating contains personally identifiable information or offensive content?
    % \answerTODO{}
    \answerNo{Data used in this paper has been open source for research purpose.}
\end{enumerate}

\item If you used crowdsourcing or conducted research with human subjects...
\begin{enumerate}
  \item Did you include the full text of instructions given to participants and screenshots, if applicable?
    % \answerTODO{}
    \answerNA{}
  \item Did you describe any potential participant risks, with links to Institutional Review Board (IRB) approvals, if applicable?
    % \answerTODO{}
    \answerNA{}
  \item Did you include the estimated hourly wage paid to participants and the total amount spent on participant compensation?
    % \answerTODO{}
    \answerNA{}
\end{enumerate}

\end{enumerate}

%%%%%%%%%%%%%%%%%%%%%%%%%%%%%%%%%%%%%%%%%%%%%%%%%%%%%%%%%%%%

\newpage
\section*{Appendix of ``Semi-supervised Semantic Segmentation with Prototype-based Consistency Regularization''}
\appendix

% The \author macro works with any number of authors. There are two commands
% used to separate the names and addresses of multiple authors: \And and \AND.
%
% Using \And between authors leaves it to LaTeX to determine where to break the
% lines. Using \AND forces a line break at that point. So, if LaTeX puts 3 of 4
% authors names on the first line, and the last on the second line, try using
% \AND instead of \And before the third author name.

% \author{%
%   David S.~Hippocampus\thanks{Use footnote for providing further information
%     about author (webpage, alternative address)---\emph{not} for acknowledging
%     funding agencies.} \\
%   Department of Computer Science\\
%   Cranberry-Lemon University\\
%   Pittsburgh, PA 15213 \\
%   \texttt{hippo@cs.cranberry-lemon.edu} \\
  % examples of more authors
  % \And
  % Coauthor \\
  % Affiliation \\
  % Address \\
  % \texttt{email} \\
  % \AND
  % Coauthor \\
  % Affiliation \\
  % Address \\
  % \texttt{email} \\
  % \And
  % Coauthor \\
  % Affiliation \\
  % Address \\
  % \texttt{email} \\
  % \And
  % Coauthor \\
  % Affiliation \\
  % Address \\
  % \texttt{email} \\
% }
% \def\lingqiao{\textcolor{red}}

% \begin{document}
% \maketitle
%%%%%%%%%%%%%%%%%%%%%%%%%%%%%%%%%%%%%%%%%%%%%%%%%%%%%%%%%%%%

% \appendix
In this appendix, we first present 
quantitative metrics for comparing the intra-/inter-class discrimination of various methods. 
Next, we provide another two ablation studies to further inspect our approach. Finally, we further visualize the semantic segmentation results of our approach for better understanding.

\section{Comparing of Intra-/Inter-class Discrimination}
In the main paper, the visualization of feature distribution in Figure 2 (c) has demonstrated that our approach can encourage a more compact within-class feature distribution and thus ease the large intra-class variation problem in the semi-supervised semantic segmentation. In order to have quantitative comparison, we borrow the principle of linear discriminant analysis (LDA) and calculate the intra-/inter-class variance of the feature representations for each comparing methods. As the results shown in Table~\ref{tab:intra_inter_disc}, our approach has not only improved the intra-class variance but also the inter-class variance, and thus the overall discrimination.
\begin{table}[h]%\footnotesize
    \centering
    \caption{Comparison of intra-/inter-class discrimination. \texttt{var.} means the variance matrix.
    }
    \label{tab:intra_inter_disc}
    \begin{tabular}{l | c | c | c }
         \toprule
         \textit{classic} VOC 2012(1/16 setting) & \makecell[c]{$\frac{tr(\text{inter-class var.})}{tr(\text{intra-class var.})}\uparrow$} & \makecell[c]{$tr(\text{inter-class var.})\uparrow$} & \makecell[c]{$tr(\text{intra-class var.})\downarrow$} \\
         \midrule
         U$^2$PL & 0.48 & 80.78 & 168.30 \\
         Ours w/o prototype-based classifier & 0.45 & 76.01 & 168.92 \\
         Ours & 2.22 & 283.43 & 127.63 \\
         \bottomrule
    \end{tabular}
\end{table}

\section{Ablation Studies}
\subsection{Strong Data Augmentation}
In the main paper, our approach is built upon the popular student-teacher weak-strong augmentation framework and the CutMix~\cite{yun2019cutmix} strong data augmentation is utilized as the default setting. In order to further investigate the effectiveness of our approach, we conduct an ablation study by varying the data augmentation approaches while keeping other modules unchanged in any comparing methods. As results shown in Table~\ref{tab:abl_strong_data_aug}, our method can still achieve overall best segmentation results with different strong data augmentations.

\begin{table}[h]%\footnotesize
    \centering
    \caption{Ablation study to the strong data augmentation on \textit{classic} PASCAL VOC 2012 1/16 setting.
    }
    \label{tab:abl_strong_data_aug}
    \begin{tabular}{l | c | c }
         \toprule
         strong data augmentation & Cutout~\cite{devries2017improved} & ClassMix~\cite{olsson2021classmix}  \\
         \midrule
         U$^2$PL & 66.82 & 67.77 \\
         Ours w/o prototype-based classifier & 66.86 & 66.93 \\
         Ours & 69.24 & 69.36 \\
         \bottomrule
    \end{tabular}
\end{table}

\subsection{Confidence Threshold}
We are also interested in how our approach will be performed when various confidence thresholds are selected. From the result shown in Table~\ref{tab:abl_confid}, we find that our approach can achieve good performance when the confidence threshold falls into a reasonable range, e.g., [0.75, 0.95]. 
\begin{table}[h]%\footnotesize
    \centering
    \caption{Ablation study of sensitivity of our approach to the selection of confidence threshold on \textit{classic} PASCAL VOC 2012 1/16 setting.
    }
    \label{tab:abl_confid}
    \begin{tabular}{l | c | c | c | c | c | c}
         \toprule
         confidence threshold & 0.95 & 0.90 & 0.85 & 0.80 & 0.75 & 0.70 \\
         \midrule
         Linear classifier & 71.01 & 70.97 & 70.30 & 70.06 & 69.43 & 64.89 \\
         Prototype-based classifier & 70.72 & 70.74 & 70.10 & 69.89 & 68.92 & 64.68 \\
         \bottomrule
    \end{tabular}
\end{table}

\section{Semantic Segmentation Visualization}
In our main paper, we have verified the effectiveness of our proposed method through extensive quantitative comparative experiments. In the appendix, we want to provide more qualitative results to further support our conclusion.

\begin{figure}[t]
\centering
\begin{center}
	\includegraphics[width=\linewidth]{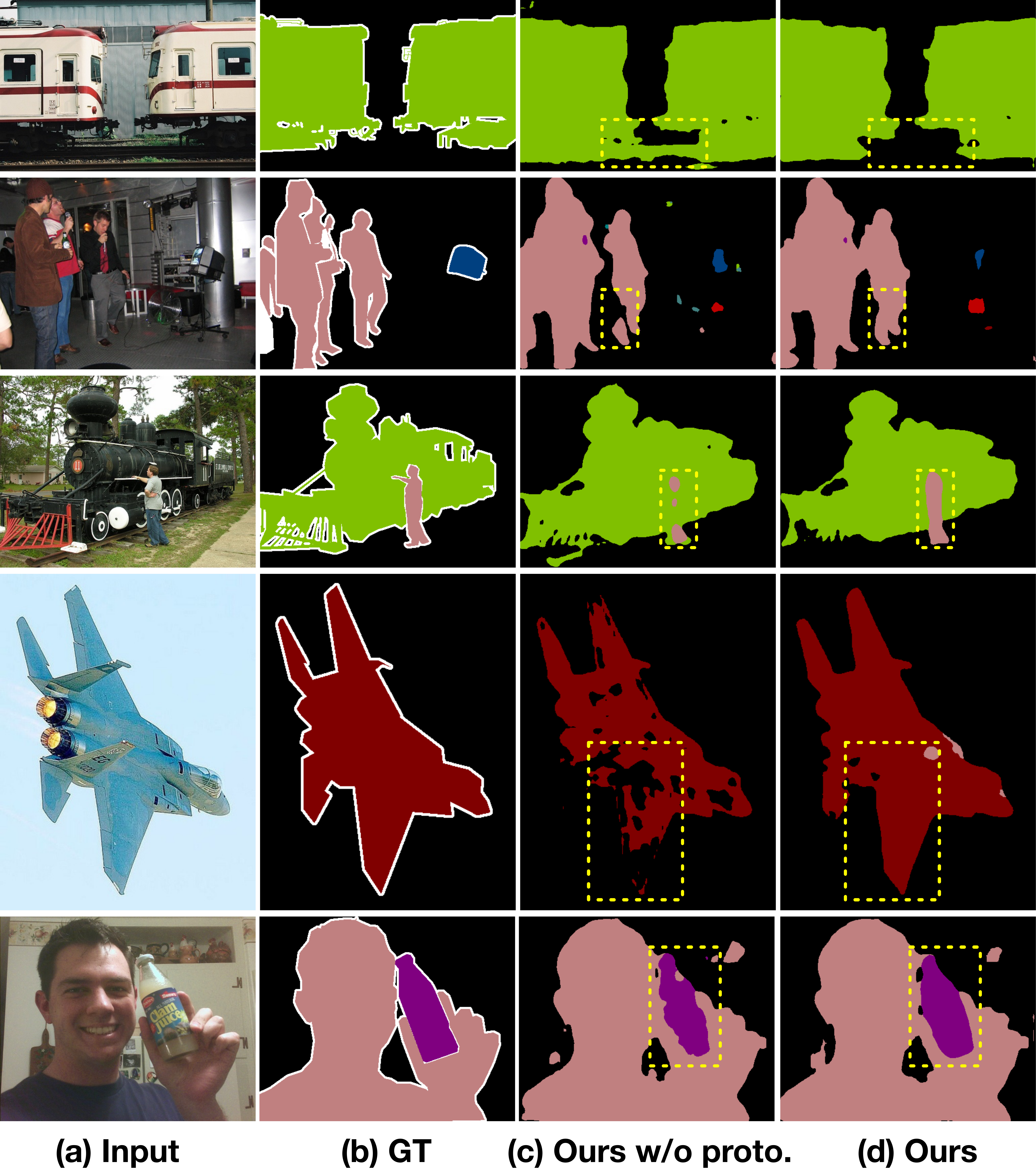}
\end{center}
\caption{Qualitative results on PASCAL VOC 2012 validation set. Methods are trained on the 1/16 label partition protocol of the \textit{classic} setting. (a) Input image, (b) Ground-truth, (c) Ours without prototype-based predictor and (d) our method. Yellow dotted boxes highlight the segments where our method performs better than the comparison method, i.e., our method can better perceive the boundary of objects.
}
\vspace{-0.5cm}
\label{fig:supp_vis_edge}
\end{figure}

\begin{figure}[t]
\centering
\begin{center}
	\includegraphics[width=\linewidth]{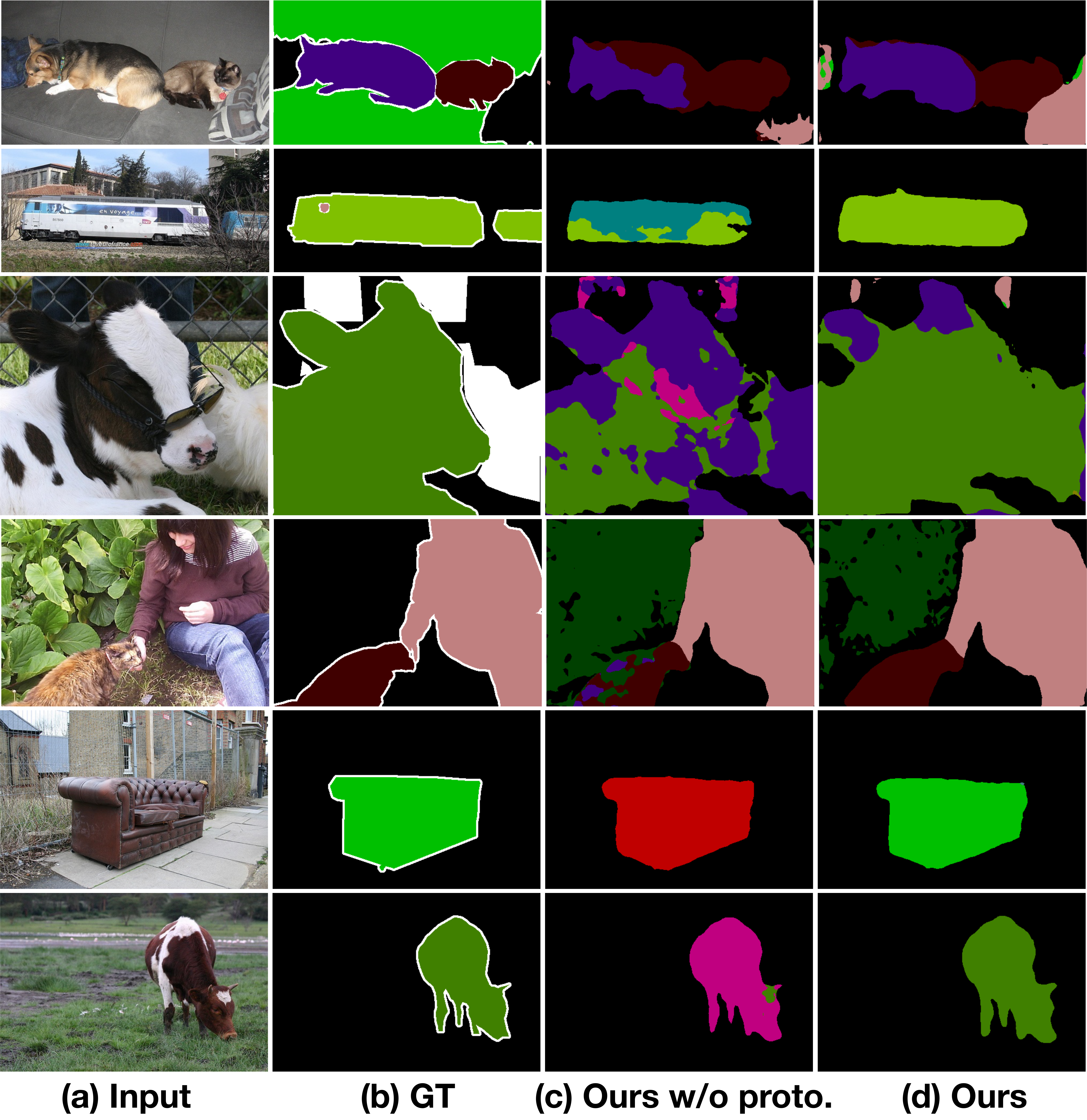}
\end{center}
\caption{Qualitative results on PASCAL VOC 2012 validation set and all methods are trained on the 1/16 label partition protocol of the \textit{classic} setting. Although both comparison methods can roughly segment the outline of the object, our approach can achieve better consistency of category prediction inner the object, especially for the objects whose appearance vary a lot, e.g., the dog at row 1 and the bus at row 2.
}
\vspace{-0.5cm}
\label{fig:supp_vis_cls}
\end{figure}

Figure~\ref{fig:supp_vis_edge} and Figure~\ref{fig:supp_vis_cls} present the segmentation results of comparing methods on the PASCAL VOC 2012 validation set from the perspective of object boundary perception and object intra-class prediction consistency, respectively. Specifically, Figure~\ref{fig:supp_vis_edge} illustrates that our method can produce better segments for the boundary of objects. As the highlighted region shown in yellow dotted boxes, i.e., the lower edge of train (row 1), the body of person (row 2-3), the wing of airplane (row 4) and the bottle (row 5), the generated segments are much more precise for our method than the baseline method without prototype-based consistency regularization constraints.

Similarly, Figure~\ref{fig:supp_vis_cls} demonstrates that our method can achieve consistent category prediction within the objects, while the comparison method may always predict different parts of the same object into different categories (e.g., the dog at row 1, the train at row 2, the cow at row 3 and the cat at row 4) and sometimes even completely wrong prediction for the whole object (the sofa at row 5 and the cow at last row are completely mispredicted as chair and horse, respectively). 

The superior semantic segmentation performance of our approach is attributed to the proposed prototype-based consistency regularization which encourages the features from the same class to be close to at least one within-class prototype while staying far away from the other between-class prototypes. Such kind of constraints will ease the label information propagation from pixels to pixels for the semi-supervised semantic segmentation task and therefore our approach can produce more precise segments and predict consistent categories within the same segment.


% Optionally include extra information (complete proofs, additional experiments and plots) in the appendix.
% This section will often be part of the supplemental material.
% \clearpage
% \bibliographystyle{plainnat}
% \bibliography{egbib}

% \end{document}

% Optionally include extra information (complete proofs, additional experiments and plots) in the appendix.
% This section will often be part of the supplemental material.

\end{document}